\documentclass{article}

\usepackage{arxiv}

\usepackage[utf8]{inputenc} 
\usepackage[T1]{fontenc}    
\usepackage{hyperref}       
\usepackage{url}            
\usepackage{booktabs}       
\usepackage{amsfonts}       
\usepackage{nicefrac}       
\usepackage{microtype}      
\usepackage{lipsum}
\usepackage{graphicx}
\usepackage{amsmath} 
\graphicspath{ {./images/} }
\usepackage{amsthm}
\newtheorem{definition}{Definition}
\usepackage{threeparttable}
\usepackage{multirow}

\title{EDSL: An Encoder-Decoder Architecture with Symbol-Level Features for Printed Mathematical Expression Recognition}

\author{
 Yingnan Fu \\
  School of Data Science and Engineering\\
  East China Normal University\\
  Shanghai, China, 200062 \\
  \texttt{yingnanfu@foxmail.com} \\
   \And
 Tingting Liu \\
  School of Data Science and Engineering\\
  East China Normal University\\
  Shanghai, China, 200062 \\
  \texttt{tingtingliu223@gmail.com} \\
  \And
 Ming Gao \\
  School of Data Science and Engineering\\
  East China Normal University\\
  Shanghai, China, 200062 \\
  \texttt{mgao@dase.ecnu.edu.cn} \\
  \And
 Aoying Zhou \\
  School of Data Science and Engineering\\
  East China Normal University\\
  Shanghai, China, 200062 \\
  \texttt{ayzhou@dase.ecnu.edu.cn} \\
}

\begin{document}
\maketitle

\begin{abstract}
	Printed Mathematical expression recognition (PMER) aims to transcribe a printed mathematical expression image into a structural expression, such as LaTeX expression. 
	It is a crucial task for many applications, including automatic question recommendation, automatic problem solving and analysis of the students, etc. 
	Currently, the mainstream solutions rely on solving image captioning tasks, all addressing image summarization.
	As such, these methods can be suboptimal for solving MER problem.
	
	In this paper, we propose a new method named EDSL, shorted for encoder-decoder with symbol-level features, to identify the printed mathematical expressions from images.
	The symbol-level image encoder of EDSL consists of segmentation module and reconstruction module.
	By performing segmentation module, we identify all the symbols and their spatial information from images in an unsupervised manner. 
	We then design a novel reconstruction module to recover the symbol dependencies after symbol segmentation.
	Especially, we employ a position correction attention mechanism to capture the spatial relationships between symbols. 
	To alleviate the negative impact from long output, we apply the transformer model for transcribing the encoded image into the sequential and structural output.
	We conduct extensive experiments on two real datasets to verify the effectiveness and rationality of our proposed EDSL method. 
	The experimental results have illustrated that EDSL has achieved 92.7\% and 89.0\% in evaluation metric Match, which are 3.47\% and 4.04\% higher than the state-of-the-art method.
	Our code and datasets are available at \url{https://github.com/abcAnonymous/EDSL}.
	
\end{abstract}

\section{Introduction}
Mathematical expression understanding is the foundation of many online education systems~\cite{liu2018fuzzy}~\cite{huang2017question}. 
It is widely used in various intelligent education applications, such as automatic question recommendation~\cite{liu2018finding}, analysis of the students~\cite{su2018exercise} and automatic problem solving~\cite{wang2018mathdqn}~\cite{wang2019template}. 
As printed math expressions often exist in the form of images, it is not conducive to analyze the formula structure and mathematical semantics. 
To understand the math expressions, it is crucial to first convert images of printed math expressions into structural expressions, such as LaTeX math expressions or symbol layout trees, which is called the printed math expression recognition, shorted in PMER.
Compared to traditional Optical Character Recognition (OCR) problem, PMER is challenging since it not only needs to identify all symbols from the images, but also captures the spatial relationships between symbols~\cite{zanibbi2012recognition}.

\begin{figure}[t]
	\begin{center}
		\includegraphics[width=0.7\linewidth]{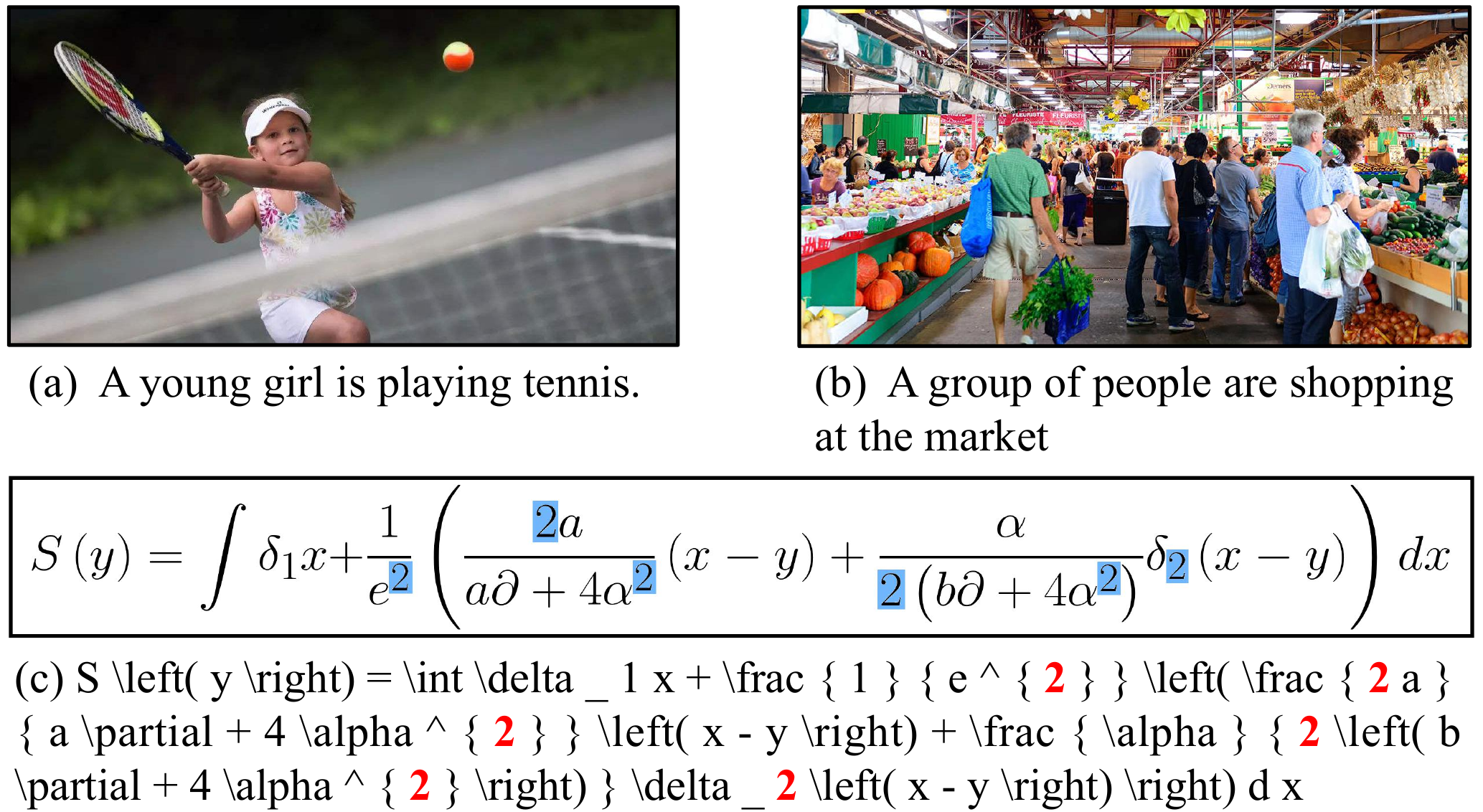}
	\end{center}
	\caption{Comparison between image captioning and PMER. (a) image captioning is insensitive to the position of the tennis ball. (b) summaries an image, rather than a detailed description. (c) illustrates that the same symbol with different positions has different mathematical semantics.}
	\label{fig:Example}
\end{figure}

OCR is most widely used to recognize math expression from images~\cite{suzuki2003infty}. 
However, traditional OCR can be significantly improved since they requires a large number of layout rules to be manually defined. 
Recent advances in PMER have focused on employing the encoder-decoder neural networks to address the image captioning problem~\cite{ren2015faster}~\cite{redmon2018yolov3}~\cite{lu2018neural}. 
In particular, encoder extracts semantic embeddings from an entire math expression image based on a convolutional neural network (CNN), and decoder predicts the LaTeX tokens using a recurrent neural network (RNN)~\cite{deng2017image}~\cite{yin2018transcribing}. 
Despite effectiveness and prevalence, we argue that these methods can be suboptimal for addressing PMER problem due to the following factors:

\begin{itemize}
	\setlength{\itemsep}{0pt} 
	\setlength{\parsep}{0pt}  
	\setlength{\parskip}{0pt}
	\item \textbf{Output sequence of PMER is Longer.} 
	The math expression is much longer than the caption of an image. 
	In the MS COCO dataset, the average length of captions is only 10.47~\cite{lin2014microsoft}. 
	However, the average length of math expressions in academic papers is 62.78~\cite{deng2017image}.
	Cho et al. demonstrate that the performance of the encoder-decoder network for image captioning deteriorates rapidly with the increase of sentence length~\cite{cho2014properties}. 
	We argue that fine-grained symbol-level features and their spatial information could alleviate the deterioration of longer outputs.
	
	
	\item \textbf{Symbol spatial information in a formula is sensitive.} 
	In a math expression, a symbol could have different mathematical semantics. 
	For a math expression illustrated in Figure~\ref{fig:Example}(c), there are six number '2', whose positions could be subscript, superscript, above and below, etc. 
	In contrast, the image captioning may be independent on the position of an object. 
	As demonstrated in Figure~\ref{fig:Example}(a), no matter where the tennis ball is, the semantics of this image will be the same. 
	%
	The image captioning models may degrade its performance for addressing MER problem since they are insensitive to capture the spatial information of objects. 
	
	\item \textbf{PMER needs to provide fine-grained description of a math expression.} 
	Image captioning aims to provide an information summarization, rather than a comprehensive, fine-grained description. 
	As demonstrated in Figure~\ref{fig:Example}(b), although there are many objects in the image, the caption still only summaries the main content in the short sentence ``\emph{A group of people are shopping at the market.}''
	However, PMER not only identifies all Roman letters, Greek letters, and operator symbols, but also needs to conduct the layout analysis of all symbols. 
	We point out that the failure of image captioning is due to the improper design for text summarization, which is suboptimal for addressing PMER problem.
\end{itemize}

Recently, the IM2Markup model proposed by Deng et al. has made efforts to solve the above problems~\cite{deng2017image}, but the shortcomings of this work are twofold:
(1) They propose the coarse-to-fine attention to reduce the computational cost.
However, this attention mechanism cannot capture the fine-grained symbol features.
(2) They employ the RNN to localize the input symbols line by line from the CNN feature map.
We argue that this approach fails to recover the spatial dependencies of symbols.
As such, IM2Markup should be further improved via capturing the fine-grained symbol features and their spatial relationships.

In this paper, we propose EDSL (shorted for encoder-decoder framework with symbol-level features), which addresses the aforementioned limitations of existing PMER methods. 
EDSL adopts a symbol-level image encoder that consists of segmentation module and reconstruction module.
The segmentation module identifies both symbol features and their spatial information in a fine-grained manner.
%
%
In the reconstruction module, we employ the position correction attention (pc-attention) to recover the spatial dependencies of symbols in the encoder.
%
For the negative impact from long output concern, we apply the transformer model~\cite{Vaswani2017Attention} to transcribe the encoded image into the sequential and structural output.
The key contributions of this paper are summarized as follows: 
\begin{itemize}
	\setlength{\itemsep}{0pt} 
	\setlength{\parsep}{0pt}  
	\setlength{\parskip}{0pt}
	\item[(1)] We propose a neural encode-decoder network with symbol-level features for addressing PMER problem (Sec.~\ref{method}). 
	To the best of our knowledge, this is the first framework that is designed to integrate segmentation and reconstruction modules into neural encoder for encoding images.

	
	\item[(2)] To recover the symbol layout, we design a PC-attention mechanism to capture the spatial relationships between symbols in the encoder (Sec.~\ref{position correction transformer encoder}).
	
	\item[(3)] We have conducted extensive experiments on two real datasets. The experimental results illustrate that EDSL significantly outperforms several state-of-the-art methods (Sec.~\ref{case study}).
\end{itemize}

\section{Related Work}

Although OCR has been used in natural language recognition and many other areas, it is difficult to recognize some special symbols and accurately reconstruct their positions for PMER.
Existing PMER methods can be categorized into two groups: traditional multi-stage methods, and end-to-end approaches. 

\subsection{Multi-Stage Methods}

A multi-stage MER method can be simplified into two sub-tasks: symbol recognition~\cite{okamoto2001performance} and symbol layout analysis~\cite{blostein1997recognition}. 

The main difficulty in symbol recognition is the problems cased by touching and over-segmented characters. 
Okamoto et al.~\cite{okamoto2001performance} used a template matching method to recognize characters. 
Alternatively, the characters can be recognized by a supervised model.
Malon et al.~\cite{malon2008mathematical} and LaViola et al.~\cite{laviola2007practical} proposed SVM and ensemble boosting classifier to improve the character recognition, respectively. 
In our proposed EDSL, we only employ unsupervised method to segment symbols from images. 
Even if there exists the case of over-segmented characters, our proposed reconstruction module will recover the symbols accurately.

The most common method used in symbol layout analysis is recursive decomposition~\cite{zanibbi2012recognition}. 
Specifically, operator-driven decomposition recursively decomposes a math expression by using operator dominance to recursively identify operators~\cite{chan2001error}. 
Projection profile cutting recursively decomposes a typeset math expression using a method similar to X-Y cutting~\cite{raja2006towards,shafait2008performance}. 
Baseline extraction decomposes a math expression by recursively identifying adjacent symbols from left-to-right on the main baseline of an expression~\cite{zanibbi2001baseline,zanibbi2002recognizing}.
%
%
In this paper, we propose an encoder-decoder framework with a PC-attention mechanism to preserve the spatial relationships, which has achieved the best performance compared to competitive baselines.

\subsection{End-to-end Methods}

Different from the multi-stage methods, MER can be also addressed by a neural encoder-decoder network with attention mechanism, where the encoder aims to understand the mathematical expression image, and decoder generates the LaTeX text.

Zhang et al.~\cite{zhang2017watch} used a VGG network as the encoder to recognize handwritten formulas. 
To improve the accuracy of handwritten formula recognition, Zhang and Du proposed a multi-scale attention mechanism based on a DenseNet network~\cite{zhang2018multi}. 
Deng et al.~\cite{deng2017image} proposed the IM2Markup model based on the coarse-to-fine attention mechanism, which achieves the state-of-the-art performance. 
Yin et al.~\cite{yin2018transcribing} proposed the spotlight mechanism to recognize of the structural images, such as math formula and music.
We argue that a CNN network is hard to directly apply for encoding math expression image features since the large receptive field cannot extract the fine-grained symbol features and small receptive field is inevitable to increase the computational cost.

In addition, image captioning can be also applied for addressing the PMER problem~\cite{luong2015effective}\cite{anderson2018bottom}\cite{chen2018regularizing}. However, we argue that image captioning is suboptimal due to the improper design for text summarization.

\section{Problem Formulation and Model Overview}

In this section, we first define the PMER problem and then introduce the overall framework of our EDSL model.

\subsection{Problem Formulation}
For a printed mathematical expression $\mathbf{x}$, which is a grayscale and structural image. 
Let $\mathbf{y}= <y_1,y_2,\cdots,y_t>$ be the sequence of LaTeX text, where $y_i$ is the $i$-th token in LaTeX sequence $\mathbf{y}$, $t$ is the sequence length. 

The task of PMER aims to transcribe the printed math expression into a LaTeX text.
Formally, the PMER problem can be defined as:
\begin{definition}[PMER problem]
	Given a printed math expression image $\mathbf{x}$, the goal of PMER is to learn a mapping function $f$, which can convert image $\mathbf{x}$ into a sequence of LaTeX text $\mathbf{y}= <y_1,y_2,\cdots,y_t>$, such that rendering $\mathbf{y}$ with a LaTeX compiler is the math expression in image $\mathbf{x}$.
\end{definition}
In the definition, PMER can be treated as a structural image transcription problem, where the structural content in an image is transcribed into a sequence of LaTeX text.

\subsection{Model Overview}
\begin{figure*}
	\begin{center}
		\includegraphics[width=0.98\linewidth]{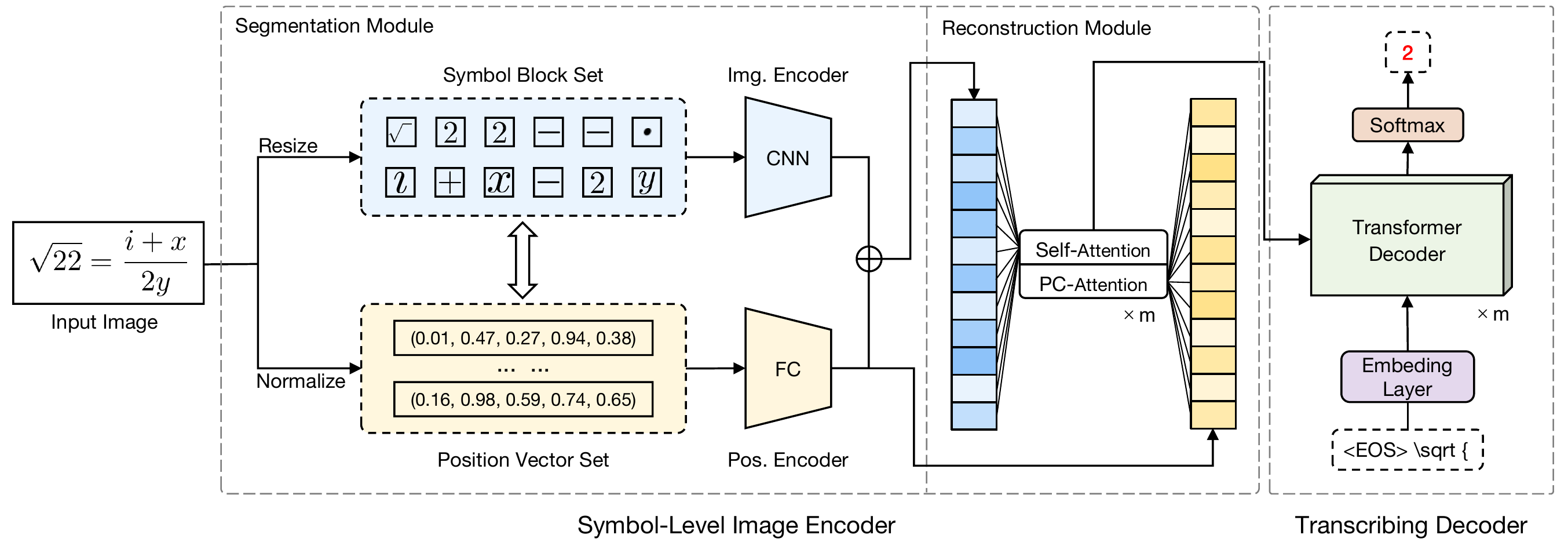}
	\end{center}
	\caption{The architecture of EDSL. EDSL consists of two main part: 1)  a symbol-level image encoder with segmentation module and reconstruction module; 2) a transcribing decoder with transformer.
	}
	\label{fig:architecture}
\end{figure*}

Figure~\ref{fig:architecture} demonstrates the overall architecture of our EDSL, which consists of two main components:
(1) \textbf{symbol-level image encoder};
(2) \textbf{transcribing decoder}. 
The encoder consists of two modules, and is designed to capture the fine-grained symbol features and their spatial information. 
The segmentation module divides an entire math expression image into symbol blocks in an unsupervised manner, such that each symbol block contains part of a symbol in the printed math expression. 
The reconstruction module is designed for recovering the spatial relationships between symbols via employing the PC-attention. 
To recover the expression, the transcribing decoder is designed to transcribe an encoding math expression into a LaTeX sequence. 

\section{Encoder-Decoder with Symbol-Level Features} \label{method}

In this section, we will explain how each part of the EDSL works in detail.

\subsection{Segmentation Module}
\begin{figure}
	\centering
	\includegraphics[width=0.8\linewidth]{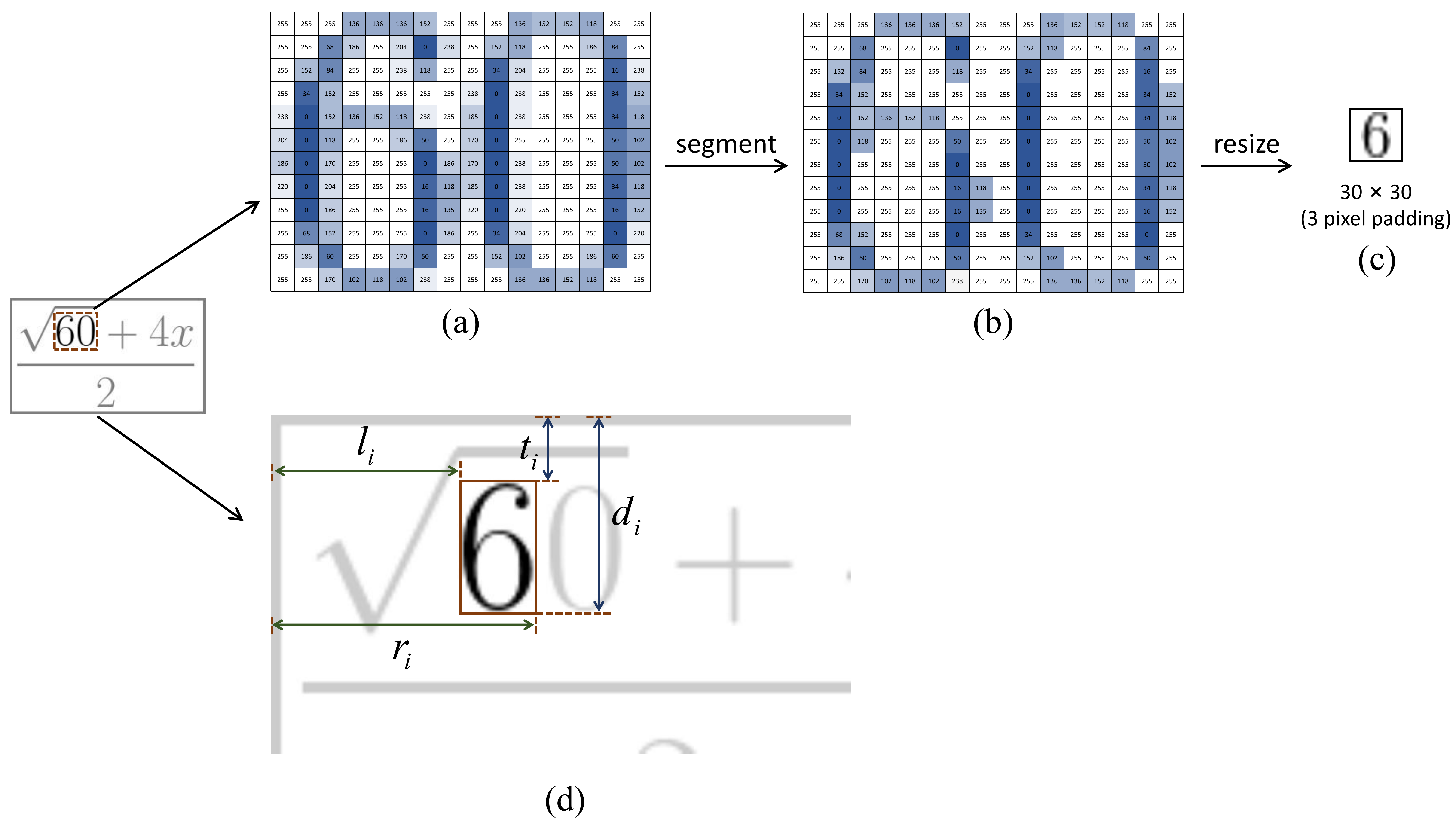}
	\caption{A running example of segmentation. (a) is a grayscale image. A connected-component labeling algorithm is used to segment (a) into two components shown in (b). (c) is the symbol block by resizing the component of (b). (d) is a diagram of position vector features of a symbol block. $t_i$, $d_i$, $l_i$, and $r_i$ are the distances from each edge (top, bottom, left, right) of the block to the upper and left of an entire image.
	}
	\label{fig:segmentation}
\end{figure}

To get the symbol-level features of a math expression, we segment an input image into symbol blocks, where each symbol block contains part of a symbol.
Given an image $\mathbf{x}$, let $S=\{\mathbf{s}_1,\mathbf{s}_2,\cdots,\mathbf{s}_n\}$ be a set of symbol blocks, where $n$ is the total number of symbol blocks, and $\mathbf{s}_i\in \mathbb{R}^{30 \times 30}$. 

It should be noted that, unlike the character segmentation used for addressing traditional OCR problem, EDSL does not need to correctly and completely segment all symbols in the image.
Each symbol block can be a complete symbol or part of a symbol, which is only utilized to extract features, rather than recognize symbols.
As such, error propagation, which commonly arises in traditional OCR task, will not take place in our proposed EDSL.
By calculating the position vectors of all symbol blocks, the spatial information of each symbol block can be preserved.
EDSL can obtain all the details in the math expression image in an unsupervised manner.
For this reason, we apply the connected-component labeling algorithm for finding symbol blocks in a math expression image~\cite{samet1988efficient}.

%

As demonstrated in Figure~\ref{fig:segmentation} (a), numbers ``6'' and ``0'' are touched symbols. 
We employ the connected-component labeling algorithm with pre-defined threshold to segment these two numbers into two components as illustrated in Figure~\ref{fig:segmentation}(b).
Subsequently, we resize each component to $30 \times 30$ pixels as illustrated in Figure~\ref{fig:segmentation} (c), where each $30 \times 30$ pixels is a symbol block.

Correspondingly, we calculate a set of position vectors $P=\{\mathbf{p}_1,\mathbf{p}_2,\cdots,\mathbf{p}_n\}$ associated with $S$, where $\mathbf{p}_i$ is the position vector of the $i-$th symbol block, and $\mathbf{p}_i = (t_i,d_i,l_i,r_i,ro_i)\in \mathbb{R}^{5}$ as illustrated in Figure~\ref{fig:segmentation}(d).
Specifically, $t_i$, $d_i$, $l_i$, and $r_i$ are the distances from each edge (top, bottom, left, right) of the $i-$th symbol block to the upper and left of input image $\mathbf{x}$. 
For ease of training, we standardize each entry of the position vector into 0 to 1 with $d_{max} = \max\{d_1,..., d_n\}$, $r_{max} = \max\{r_1, ..., r_n\}$ as follows:
\begin{equation}
	\mathbf{p}_i = (\frac{t_i}{d_{max}}, \frac{d_i}{d_{max}},\frac{l_i}{r_{max}},\frac{r_i}{r_{max}},\frac{d_{max}}{r_{max}}).
\end{equation} 
Astute readers may find the last entry $\frac{d_{max}}{r_{max}} = ro_i$, which is the width/height ratio of input image $\mathbf{x}$, and will help us to reconstruct a symbol when it is distorted after standardizing the first 4 entries. 

For preserving the symbol features and spatial information, we employ image and position encoders to map each symbol block into two low-dimension spaces as demonstrated in Figure~\ref{fig:architecture}. 
Specifically, we employ a six-layer CNN model with a fully connected layer~\cite{simonyan2014very} to encode all symbol blocks of $S$ into a 256-dimension space, denoted as $S^{\prime}=\{\mathbf{s}_1^{\prime},\mathbf{s}_2^{\prime}, \cdots, \mathbf{s}_n^{\prime}\}$, where $\mathbf{s}_i^{\prime} \in \mathbb{R}^{256}$.
Similarly, we employ a three-layer fully connected network to encode all position vectors, and also embed them into a 256-dimension space, denoted as 
$P^{\prime}= \{\mathbf{p}_1^{\prime},\mathbf{p}_2^{\prime}, \cdots, \mathbf{p}_n^{\prime}\} $, where $\mathbf{p}_i^{\prime} \in \mathbb{R}^{256}$.
%

Finally, we can get the symbol block embedding set 
\begin{equation}
	E=\{\mathbf{e}_1,\mathbf{e}_2,\cdots, \mathbf{e}_n\}
\end{equation}
where 
\begin{equation}
	\mathbf{e}_i = \mathbf{s}_i^{\prime} + \mathbf{p}_i^{\prime},\mbox{\;\;\;\; for } i \in 1, \cdots, n.
\end{equation}

\subsection{Reconstruction Module} \label{position correction transformer encoder}
Since the embedding vectors of all symbols are independent, it is necessary to reconstruct the spatial relationships between symbols.
Although RNN is common used approach to infer the dependencies between entries in a sequence, symbol blocks are in a two-dimensional space and cannot be modeled as a sequence.
To reconstruct  the spatial relationships between symbols, we employ a transformer model with a novel attention mechanism.
The symbol block embedding set $E$ is encoded to embedding set $R = \{\mathbf{r}_1,\mathbf{r}_2,...,\mathbf{r}_n\}$.
Now we discuss self-attention and our proposed position correction attention(PC-attention).


\textbf{Self-Attention.} 
It refers to calculating attention score within a sequence, which can learn the dependencies between tokens.
For each symbol block vector in $E$, we need to calculate the attention scores with all other symbols.
As such, we can capture the internal spatial relationships between symbol blocks.

The attention score between each pair of symbol block vectors is calculated by scaled dot-product attention:
\begin{equation}
	\begin{aligned}
		&    \alpha(\mathbf{e}_i,\mathbf{e}_j) = \frac{\mathbf{e}_i^{T} \cdot \mathbf{e}_j}{\sqrt{d_e}}  \\
		&    \textup{attn}(\mathbf{e}_i,\mathbf{e}_j) = \frac{\exp{(\alpha(\mathbf{e}_i,\mathbf{e}_j))}}{\sum_{k=1}^n\exp{(\alpha(\mathbf{e}_i,\mathbf{e}_k))}}
	\end{aligned}
	\label{self-attention}
\end{equation} 
where $d_e$ is the dimension of $\mathbf{e}_i$.
In practice, we utilize the multi-head variant to calculate $\textup{attn}(\mathbf{e}_i,\mathbf{e}_j)$~\cite{Vaswani2017Attention}.

\textbf{PC-Attention.}
Since self-attention is a global attention mechanism, it may be suboptimal to capture the spatial relationships in a long math expression since a symbol is not necessary to interact with the other symbols far away from it.
Thus, we introduce the position correction attention (PC-attention), which utilizes the position vectors $\mathbf{p}_i^{\prime}$ to calculate attention scores for a target symbol. 

For each pair of symbol block vectors, PC-attention first calculates the attention score $\alpha_{pos}$ of their symbol block vectors followed by~\cite{luong2015effective}.
Then add it with $\alpha$ to obtain the new attention weight $\alpha^{\prime}$. 
Finally PC-attention score $\mathrm{attn}^{\prime}$ can be calculated by normalized $\alpha^{\prime}$ by with softmax function.
The PC-attention score is calculated as follows:
\begin{equation}
	\begin{aligned}
		&    \alpha_{pos}(\mathbf{p}_i^{\prime}, \mathbf{p}_j^{\prime}) = \mathbf{v}^T_a \tanh(\mathbf{W}_a[\mathbf{p}_i^{\prime};\mathbf{p}_j^{\prime}])  \\
		&    \alpha ^ {\prime}(\mathbf{e}_i,\mathbf{e}_j) =  \alpha(\mathbf{e}_i, \mathbf{e}_j) + \alpha_{pos}(\mathbf{p}_i^{\prime}, \mathbf{p}_j^{\prime}) \\  
		&    \textup{attn}^{\prime}(\mathbf{e}_i,\mathbf{e}_j) = \frac{\exp{(\alpha^{\prime}(\mathbf{e}_i,\mathbf{e}_j))}}{\sum_{k=1}^n\exp{(\alpha^{\prime}(\mathbf{e}_i,\mathbf{e}_k))}}
	\end{aligned}
\end{equation} 
where $\mathbf{W}_a \in \mathbb{R} ^ {512 \times 256}$, $\mathbf{v}_a \in \mathbb{R} ^ {256}$.
We also utilize the multi-head variant to calculate $\textup{attn}^{\prime}(\mathbf{e}_i,\mathbf{e}_j)$ in practice.

\begin{figure}
	\centering
	\includegraphics[width=0.6\linewidth]{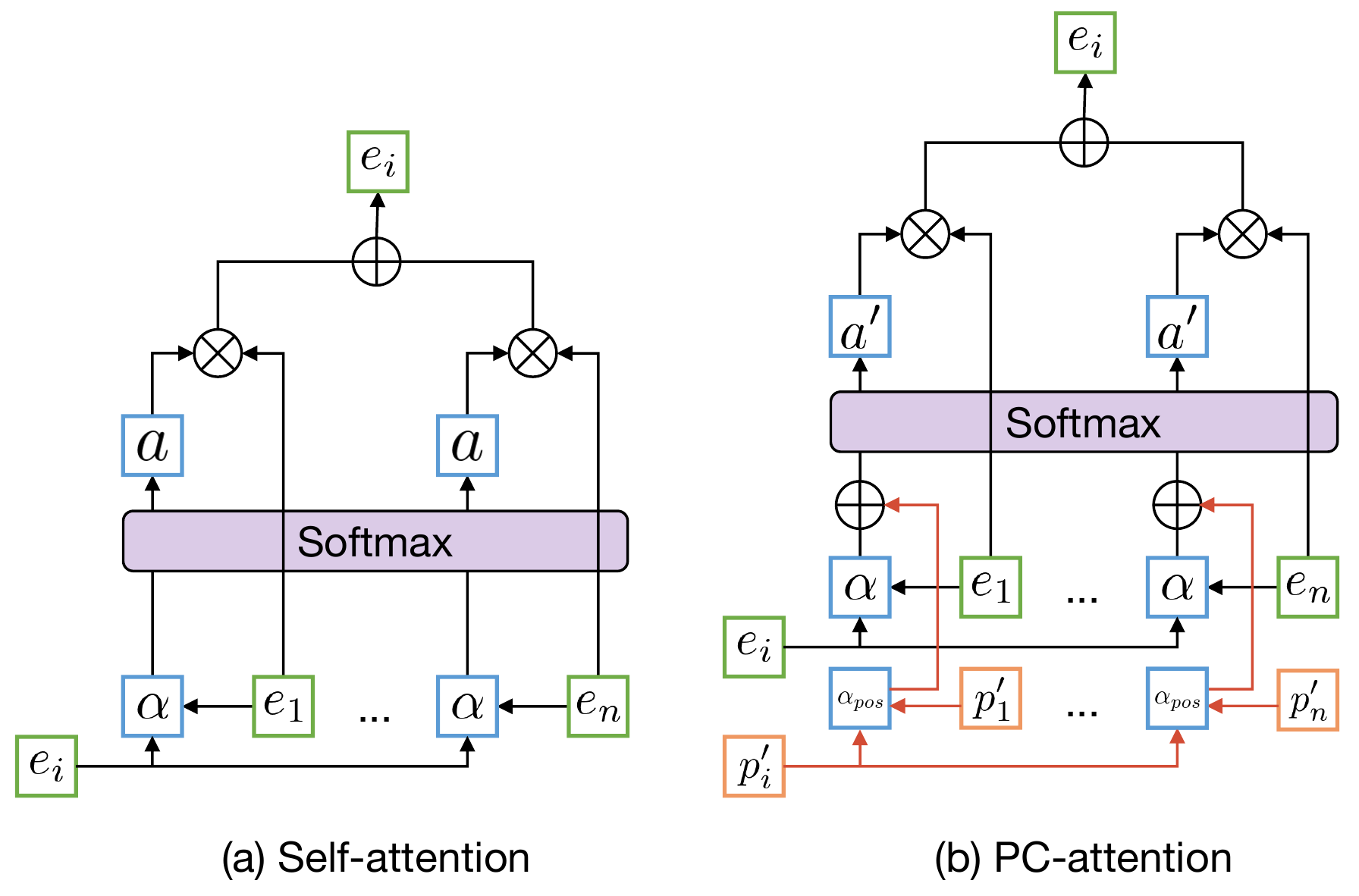}
	
	\caption{Comparison of self-attention and PC-attention.
	}
	\label{fig:attention}
\end{figure}

The comparison of self-attention and PC-attention is illustrated in Figure~\ref{fig:attention}.
Although PC-attention appears to be more complex, $pi^{\prime}$ is unnecessary to be updated during the calculation. 
As such, PC-attention is as efficient as self-attention since $\alpha_{pos}$ only need to be calculated once.

PC-attention calculates the attention score via combining both symbol features and their spatial information.
To avoid unnecessary long-distance dependencies, PC-attention focuses on the nearest symbols via adjusting self-attention with position information to better reconstruct the spatial relationships between symbols.

\subsection{Transcribing Decoder}

As with the general encoder-decoder architecture, the transcribing decoder of EDSL generates one token at each time by given the symbol block embeddings set $R$ and previous outputs.
We employ a transformer model as decoder to transcribe the math expression since it is more conducive to generate long LaTeX sequence compared with others.
The multi-head attention followed by~\cite{Vaswani2017Attention} is employed between symbol block embeddings and transcribing decoder. 
To build the decoder, we define the following language model on the top of the transformer:
\begin{equation}
	p(y_{t}|y_1, ..., y_{t-1}, \mathbf{r}_1, ..., \mathbf{r}_n) = \textup{softmax}(\mathbf{W}_{out} \mathbf{o}_{t-1})
\end{equation} 
where $y_{t}$ is $t-$th token in the output LaTeX sequence, $\mathbf{o}_{t-1}$ is output of the transformer decoder in the $(t-1)$th step, $\mathbf{W}_{out} \in \mathbb{R}^{256 \times |v|}$, and $|v|$ is the vocabulary size.
The overall loss $\mathcal{L}$ is defined as the negative log-likelihood of the LaTeX token sequence:
\begin{equation}
	\mathcal{L} = \sum_{t=1}^{T} - \log P(y_t|y_1,...,y_{t-1},\mathbf{r}_1,...,\mathbf{r}_n)
\end{equation} 
Since all calculations are deterministic and differentiable, the model can be optimized by standard back-propagation.

\section{Experiment}

To evaluate the performance of EDSL, we conduct extensive experiments on two real datasets. 
Through empirical studies, we aim to answer the following research questions:
\begin{enumerate}
	\setlength{\itemsep}{0pt} 
	\setlength{\parsep}{0pt}  
	\setlength{\parskip}{0pt}
	\item[\bf RQ1:] How does EDSL perform compared with the state-of-the-art methods and other representative baselines?
	
	\item[\bf RQ2:] How does the length of math expression affect the performance of EDSL?
	
	\item[\bf RQ3:] Is symbol-level image encoder helpful to improve the performance of EDSL?
\end{enumerate}
In addition, we conduct a case study, which visualizes the role of different attention mechanisms.

\subsection{Experimental Setup}
\subsubsection{Dataset}
We evaluate the performance of EDSL on two public datasets, Formula~\cite{yin2018transcribing} and IM2LATEX~\cite{deng2017image}.
Before reporting the performance, we pre-process the two datasets as follows: 

\textbf{ME-20K:} Dataset Formula collects printed math expression images and corresponding LaTeX representations from high school math exercises in Zhixue.com, which is an online education system.
Due to many duplicates existed in the dataset, we remove the duplicates and rename the new dataset as ME-20K.

\textbf{ME-98K:} Dataset IM2LATEX collects the printed formula and corresponding LaTeX representations from 60,000 research papers. 
As there are 4881 instances in the IM2LATEX dataset, which are tables or graphs, rather than math expressions, We remove these LaTeX strings and corresponding images from IM2LATEX, and get the dataset named ME-98K.

The statistics of our experimental datasets are summarized in Table ~\ref{Dataset}. 
We can observe that ME-20K is short in length and simple in structure.

\begin{table}[h]
	\caption{Statistics of ME-20K and ME-98K.}
	\label{Dataset}
	\centering
	\begin{tabular}{ccccc}
		\toprule
		Dataset & \begin{tabular}[c]{@{}c@{}}Image count\end{tabular} & \begin{tabular}[c]{@{}c@{}}Token space\end{tabular} & \begin{tabular}[c]{@{}c@{}}Avg. tokens per images\end{tabular} & \begin{tabular}[c]{@{}c@{}}Max tokens per image\end{tabular} \\ 
		\midrule
		ME-20K  & 20834                                                 & 191                                                   & 16.27                                                            & 181                                                            \\ 
		ME-98K  & 98676                                                 & 297                                                   & 62.78                                                            & 1050                                                           \\ 
		\toprule
	\end{tabular}
\end{table}

\subsubsection{Baselines}
We compare EDSL with two types of baselines:

\textbf{MER Method.} IM2Markup employs an encoder-decoder model with coarse-to-fine attention for recognizing math expressions~\cite{deng2017image}. It has achieved state-of-the-art performance on dataset IM2LATEX. Due to poor performance reported in~\cite{deng2017image}, we do not report the performance of other PMER methods, such as INFTY~\cite{suzuki2003infty} and CTC~\cite{shi2016end}.

\textbf{Image Captioning Methods.} We also compare our EDSL with several competitive image captioning methods.
\begin{itemize}
	\setlength{\itemsep}{0pt} 
	\setlength{\parsep}{0pt}  
	\setlength{\parskip}{0pt}
	\item SAT~\cite{xu2015show}: This method proposes a soft-attention mechanism, which uses a fully connected network to calculate the attention scores.
	
	\item DA~\cite{luong2015effective}: This method utilizes the dot-product attention mechanism to address the image captioning problem.
	
	\item TopDown~\cite{anderson2018bottom}: This method proposes a Top-Down attention mechanism, which uses two LSTM layers to selectively focus on spatial image features and predict the current output.
	
	\item ARNet~\cite{chen2018regularizing}: This method regularizes the transition dynamics of recurrent neural networks and further improves the performance of language model for image captioning.

	\item LBPF~\cite{qin2019look}: This method combines past and future information to improve the performance of image captioning.
	
	\item CIC~\cite{aneja2018convolutional}: This method only uses a convolutional language model as the decoder.
\end{itemize}

Moreover, we propose two implementations on the basis of our proposed EDSL method. The first one is EDSL-S, which employs the self-attention mechanism to capture the spatial relationships between symbols. Comparatively, the second is a more sophisticated one, i.e., EDSL-P with position correction attention (PC-attention).

\subsubsection{Evaluation Metrics}
Our main evaluation method is to check the matching accuracy of the rendered prediction image compared to the ground-truth image. 
Followed by ~\cite{deng2017image}, we also employ Match-ws to check the exact match accuracy after eliminating white space columns. 
Besides, we also use standard text generation metrics, BLEU-4~\cite{papineni2002bleu} and ROUGE-4~\cite{lin2004rouge}, to measure the precision and recall of the tokens in output sequences. 
All experiments are conducted three times and a paired t-test is performed on each metric to ensure the significance of the experimental results. 
$^{*}$ and $^{**}$ indicate that the improvements are statistically significant for $p<0.05$ and $p<0.01$ judged by paired t-test, respectively.

\subsubsection{Implementation Details}
As mentioned in ~\cite{deng2017image},  we group the images into similar sizes to facilitate batching for baselines~\footnote {Width-height group sizes are (128,32), (128,64), (160,32), (160,64), (192,32), (192,64), (224,32), (224,64), (256,32), (256,64), (320,32), (320,64), (384,32), (384,64), (384,96), (480,32), (480,64), (480,128), (480,160).}. 
In EDSL, we employ two 8-layer transformer models with eight heads as reconstruction module and transcribing decoder. 
The embedding size of EDSL is 256. 
We also use 160, 180, 200 as the segmentation thresholds on the training set and keep different symbol blocks of the same image as different training samples. 
In this way, the training samples roughly tripled.
The default threshold is 160 for both validation set and test set.
We make this approach as data augmentation for training the EDSL model. 
The effect of different thresholds for segmentation is further discussed in Sec.~\ref{character segmentation discussion}.

We train our models on the GTX 1080Ti GPU. The batch size of ME-20K and ME-98K are 32 and 16, respectively. We use Adam optimizer with an initial learning rate of 0.0003. Once the validation loss does not decrease in three epochs, we halve the learning rate. We stop training if it does not decrease in ten epochs.

\subsection{Performance Comparison(RQ1)} \label{general performance}

\begin{table}
	\caption{Performance comparison on ME-20K and ME-98K. }
	\label{tabel:baselines match}
	\centering
	\setlength{\tabcolsep}{3mm}{
		\begin{tabular}{c|cccccc}
			\toprule
			Dataset                 & Type                  & Method          & Match-ws       & Match          & BLEU-4         & ROUGE-4        \\ \midrule
			\multirow{9}{*}{ME-20K} & \multirow{6}{*}{I.C.} & CIC             & 70.91          & 70.56          & 79.27          & 83.37          \\
			&                       & DA              & 77.31          & 76.92          & 87.27          & 89.08          \\
			&                       & LBPF            & 80.88          & 80.46          & 88.82          & 90.57          \\
			&                       & SAT             & 82.65          & 82.09          & 89.77          & 91.15          \\
			&                       & TopDown         & 84.22          & 83.85          & 90.55          & 91.94          \\
			&                       & ARNet           & 85.84          & 85.40          & 91.18          & 92.50          \\ \cline{2-7} 
			& \multirow{3}{*}{MER}  & IM2Markup       & 89.63          & 89.23          & 92.83          & 93.74          \\
			&                       & EDSL-S          & 92.39          & 91.55          & 93.91          & 94.77          \\
			&                       & \textbf{EDSL-P} & $\mathbf{93.45}^{**}$ & $\mathbf{92.70}^{**}$ & $\mathbf{94.23}$ & $\mathbf{95.10}$ \\ \midrule
			\multirow{9}{*}{ME-98K} & \multirow{6}{*}{I.C.} & CIC             & 33.71          & 33.62          & 55.47          & 65.52          \\
			&                       & DA              & 55.15          & 55.15          & 79.71          & 82.40          \\
			&                       & LBPF            & 66.87          & 66.83          & 84.64          & 86.57          \\
			&                       & SAT             & 71.04          & 70.85          & 86.56          & 87.86          \\
			&                       & TopDown         & 72.85          & 72.65          & 87.56          & 89.32          \\
			&                       & ARNet           & 68.98          & 68.55          & 86.04          & 88.27          \\ \cline{2-7} 
			& \multirow{3}{*}{MER}  & IM2Markup       & 85.16          & 84.96          & 91.47          & 92.45          \\
			&                       & EDSL-S          & 88.02          & 87.50          & 92.65          & 93.08          \\
			&                       & \textbf{EDSL-P} & $\mathbf{89.34}^{**}$ & $\mathbf{89.00}^{**}$ & $\mathbf{92.93}$ & $\mathbf{93.30}$ \\ \toprule
	\end{tabular}}
	\begin{tablenotes}
		\footnotesize
		\centering
		\item  ** indicates that the improvements are statistically significant for $p < 0.01$ judged by paired t-test.
	\end{tablenotes}
\end{table}

\begin{table}
	\caption{Cumulative attention scores of the nearest symbols with PC-attention and self-attention.}
	\label{table:pcattention and self-attention}
	\centering
	\setlength{\tabcolsep}{7mm}{
		\begin{tabular}{ccccc}
			\toprule
			\multirow{2}{*}{Dataset} & \multirow{2}{*}{Method} & \multicolumn{3}{c}{Cumulative Attention Scores} \\ \cline{3-5} 
			&                         & 10\%           & 20\%           & 30\%          \\ \midrule
			\multirow{2}{*}{ME-20K}  & EDLS-S                  & 0.130          & 0.316          & 0.435         \\
			& EDSL-P                  & $\mathbf{0.261}^{**}$        & $\mathbf{0.514}^{**}$        & $\mathbf{0.636}^{**}$       \\ \midrule
			\multirow{2}{*}{ME-98K}  & EDLS-S                  & 0.531          & 0.656          & 0.692         \\
			& EDSL-P                  & $\mathbf{0.597}^{**}$        & $\mathbf{0.757}^{**}$        & $\mathbf{0.801}^{**}$       \\ 
			\toprule
	\end{tabular}}
	\begin{tablenotes}
		\footnotesize
		\centering
		\item  ** indicates that the improvements are statistically significant for $p < 0.01$ judged by paired t-test.
	\end{tablenotes}
\end{table}

Table~\ref{tabel:baselines match} illustrates the performance of baselines and our proposed EDSL method, where we have the following key observations:
\begin{itemize}
	\setlength{\itemsep}{0pt} 
	\setlength{\parsep}{0pt}  
	\setlength{\parskip}{0pt}
	\item MER methods outperform the image captioning baselines. This is due to the factors that: (1) image captioning methods aim to summary an input image, rather than design for mining the fine-grained spatial relationships between symbols; (2) MER methods, including IM2Markup, EDSL-S and EDSL-P, are designed to reconstruct the spatial relationships between symbols in the fine-grained manner, which are more advantageous.
	
	\item Both EDSL-S and EDSL-P approaches are significantly better than IM2Markup. This improvement illustrates the effectiveness of EDSL, which employs the symbol-level image encoder to capture both symbol features and their spatial information, and preserves more details compared with IM2Markup.
	
	\item EDSL-P outperforms EDSL-S and achieves the best performance on both datasets. This points to the positive effect of employing PC-attention mechanism to reconstruct the spatial relationships between symbols in the image encoder.
\end{itemize}

To better understand the mechanism of PC-attention and self-attention, we further calculate the cumulative attention scores of the top-$k$\% nearest symbols for a target character. 
We report the average cumulative attention score for all symbols in Table~\ref{table:pcattention and self-attention}. 
Since longer math expression has a higher cumulative attention score, the average cumulative attention score is higher in ME-98K dataset.
We can observe that the cumulative attention score of PC-attention is higher than that of self-attention. 
It indicates that PC-attention tends to infer the local spatial dependencies to reconstruct the whole math expression. 
Actually, recovery of local spatial dependencies is crucial for EDSL after segmentation in the symbol-level encoder.

\subsection{Effect of Sequence Lengths (RQ2)}

\begin{figure}[t]
	\centering
	\includegraphics[width=0.8\linewidth]{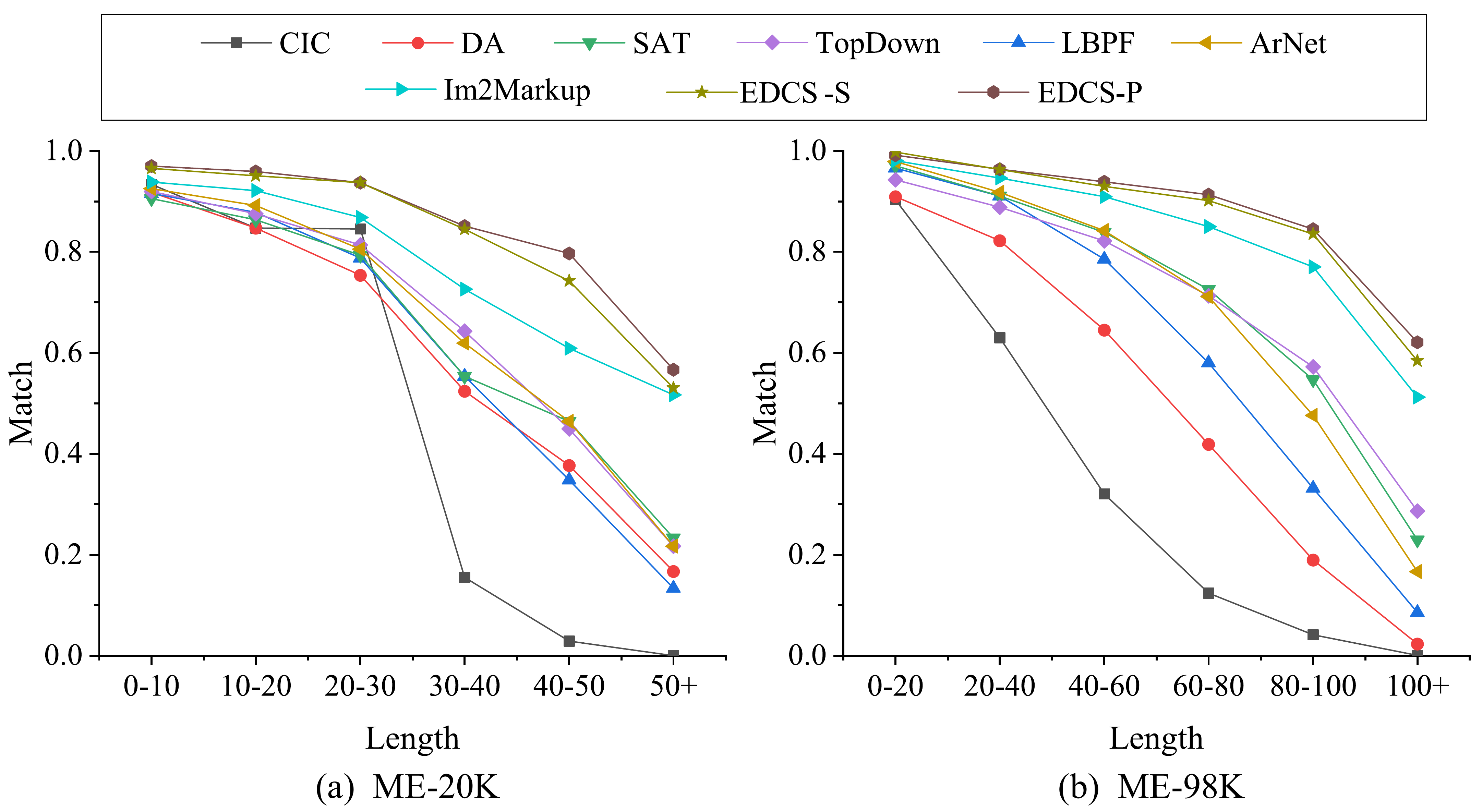}
	\caption{Performance with different math expression lengths on ME-20K and ME-98K.}
	\label{fig:ratio}
\end{figure}

To demonstrate the effect of formula lengths, we vary the match expression lengths to evaluate the performances of baselines and our proposed EDSL method. 
As illustrated in Figure~\ref{fig:ratio}, we have the following observations:

\begin{itemize}
	\setlength{\itemsep}{0pt} 
	\setlength{\parsep}{0pt}  
	\setlength{\parskip}{0pt}
	\item The length of math expression affects the performances of all methods significantly. 
	This is due to the factor that the neural encoder-decoder models will significantly decrease as the sequence length increases~\cite{cho2014properties}. 
	It indicates a negative impact on long math expressions.
	
	\item Both EDSL-S and EDSL-P methods have achieved better performances when the math expression lengths vary. 
	This sheds lights on the benefit of preserving the fine-grained symbol-level features and their spatial information in the symbol-level image encoder. 
	Although the performances of both EDSL-S and EDSL-P also decrease as the length of math expression increases, the performance declines are much smaller than the others.
	This indicates that EDSL is qualified to recognize the long math expressions.
\end{itemize}

\subsection{Utility of Symbol-Level Image Encoder (RQ3)}

\begin{table}
	\caption{Comparison of the performances of EDSL-S, EDSL-P and its variant methods ED and ED + Seg.}
	\label{table:segmentation}
	\centering
	\begin{threeparttable}
		\setlength{\tabcolsep}{4mm}{
			\begin{tabular}{cccccc}
				\toprule
				Dataset                 & Method          & Match-ws         & Match            & BLEU-4         & ROUGE-4        \\ \midrule
				\multirow{4}{*}{ME-20K} & ED              & 79.75            & 79.31            & 89.24          & 90.69          \\
				& ED+Seg          & 88.70            & 88.26            & 92.76          & 93.65          \\
				& EDSL-S          & 92.39            & 91.55            & 93.91          & 94.77          \\
				& \textbf{EDSL-P} & $\mathbf{93.45}^{**}$ & $\mathbf{92.70}^{**}$ & $\mathbf{94.23}$ & \textbf{95.10} \\ \midrule
				\multirow{4}{*}{ME-98K} & ED              & 68.71            & 68.54            & 86.15          & 87.31          \\
				& ED+Seg          & 81.15            & 80.57            & 91.58          & 91.97          \\
				& EDSL-S          & 88.02            & 87.50            & 92.65          & 93.08          \\
				& \textbf{EDSL-P} & $\mathbf{89.34}^{**}$ & $\mathbf{89.00}^{**}$ & $\mathbf{92.93}$ & $\mathbf{93.30}$ \\ \toprule
		\end{tabular}}
		\begin{tablenotes}
			\footnotesize
			\centering
			\item  ** indicates that the improvements are statistically significant for $p < 0.01$ judged by paired t-test.
		\end{tablenotes}
	\end{threeparttable}
\end{table}

To demonstrate the effectiveness of symbol-level image encoder, we compare EDSL-S and EDSL-P with their variants method ED and ED + Seg.
ED only employs CNN model as the encoder and a transformer model as the decoder, and takes the entire image of mathematical expression as input. 
ED + Seg removes the reconstruction module from the symbol-level image encoder of EDSL.
From Table~\ref{table:segmentation}, we have the following key observations:

\begin{itemize}
	\setlength{\itemsep}{0pt} 
	\setlength{\parsep}{0pt}  
	\setlength{\parskip}{0pt}
	%
	%
	
	\item Comparing ED with ED + Seg, the values of Match are improved by 8.95\% and 12.03\% on two datasets, respectively.
	This is due to the factor that ED + Seg encodes the fine-grained symbol features.
	These improvements prove the effectiveness of the fine-grained symbols features captured by the segmentation module.
	
	\item Comparing ED with ED + Seg on two datasets, the performance improvement on ME-98K is much higher.
	It reveals that our designed symbol-level image encoder has more obvious advantage on transcribing the longer math expression.
	
	\item EDSL-P outperforms the others significantly.
	This is due to the factor that PC-attention is designed for recovering the spatial relationships of symbols.
	This again points to the positive effect of employing PC-attention mechanism to reconstruct the spatial relationships between symbols in reconstruction module of encoder.
\end{itemize}

\subsection{Hyper-Parameter Studies}\label{character segmentation discussion}

\begin{table}
	\caption{EDSL-P performance of varying segmentation thresholds on both dataset, where Th is the threshold used in the segmentation algorithm.}
	\label{table:threshold}
	\centering
	\begin{threeparttable}
		\setlength{\tabcolsep}{4.8mm}{
			\begin{tabular}{cccccc}
				\toprule
				Dataset                 & Th          & Match-ws        & Match           & BLEU-4          & ROUGE-4         \\ \midrule
				\multirow{4}{*}{ME-20K} & 160         & 92.56           & 91.82           & 93.97           & 94.85           \\
				& 180         & 92.76           & 92.00           & 93.26           & 95.06           \\
				& 200         & 91.82           & 91.77           & 93.66           & 94.52           \\
				& \textbf{DA} & $\mathbf{93.45}^{*}$ & $\mathbf{92.70}^{*}$ & $\mathbf{94.34}^{*}$ & $\mathbf{95.10}^{*}$ \\ \midrule
				\multirow{4}{*}{ME-98K} & 160         & 87.35           & 87.06           & 92.75           & 93.11           \\
				& 180         & 85.53           & 85.16           & 92.39           & 92.75           \\
				& 200         & 85.72           & 85.38           & 92.35           & 92.72           \\
				& \textbf{DA} & $\mathbf{89.34}^{*}$ & $\mathbf{89.00}^{*}$ & $\mathbf{92.93}^{*}$ & $\mathbf{93.30}^{*}$ \\ \toprule
		\end{tabular}}
		\begin{tablenotes}
			\footnotesize
			\centering
			\item  * indicates that the improvements are statistically significant for $p < 0.05$ judged by paired t-test.
		\end{tablenotes}
	\end{threeparttable}
\end{table}

Different segmentation thresholds will produce different symbol blocks, which fundamentally affects the encoder to extract the symbol features and their spatial information.
We therefore investigate the impact of threshold used for segmentation.
As demonstrated in Table~\ref{table:threshold}, we vary the threshold from 160 to 200, and observe that the different segmentation thresholds do influence on the performance of EDSL. 
This is due to the factor that different segmentation thresholds will produce different symbol blocks, which affects the results of image feature extraction.

Inspired by the data augmentation, we retain segmented symbols given by different segmentation thresholds to increase the diversity of data for training EDSL, denoted as DA. 
We can observe that our EDSL method can be further improved after data augmentation. 
It indicates that we can use the diversity of segmentation results to improve the performance and avoid the difficulty of threshold selection.

\subsection{Case Study}\label{case study}

\begin{figure*}
	\centering
	\includegraphics[width=0.9\linewidth]{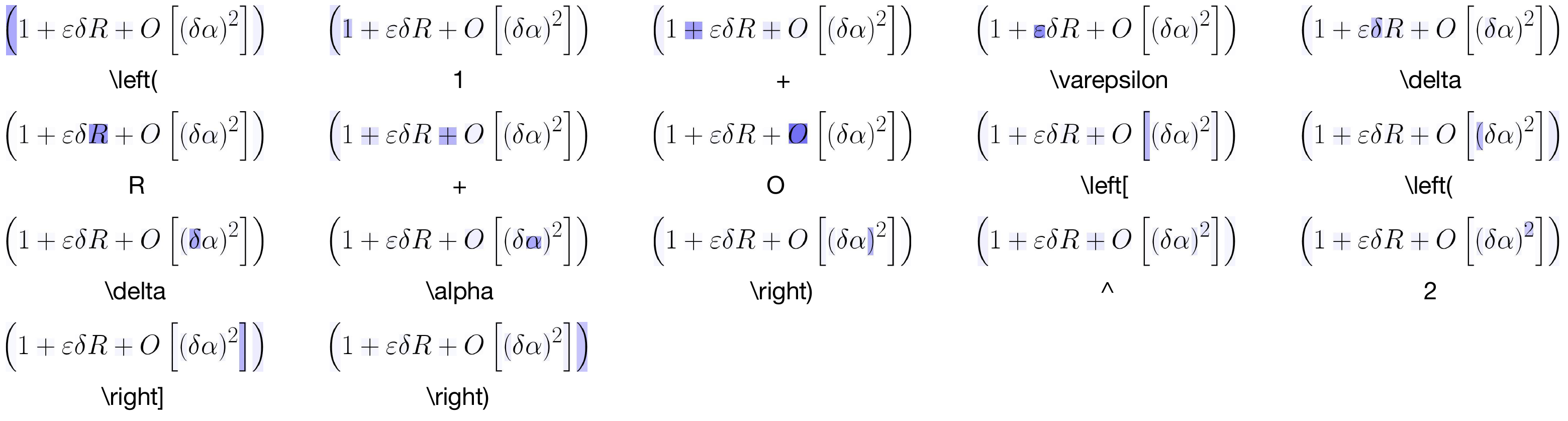}
	\caption{Visualization of predicted tokens and attention maps. The LaTeX text is "$\backslash$left( 1 + $\backslash$varepsilon $\backslash$delta R + O $\backslash$left[ $\backslash$left( $\backslash$delta $\backslash$alpha $\backslash$right) $\bigwedge$ 2 $\backslash$right]  $\backslash$right). Darker color means a larger attention weight."}
	\label{fig:Decoder}
\end{figure*}

To better understand our proposed EDSL model, we visualize the attention scores for the tokens in the output LaTex text. 
We fetch the attention scores in the last layer of the transcribing decoder. 
Figure~\ref{fig:Decoder} demonstrates the predict tokens and the attention map. 
We can observe that: (1) For an output token, EDSL only focuses on the whole corresponding symbols, rather than a region given by image captioning methods~\cite{xu2015show,lu2017knowing}; (2) even if there are many identical symbols in an math expression image, EDSL is able to focus on the correct position.
These shed lights on the benefit of symbol-level image encoder, which is helpful to recognize all symbols and their spatial information.

\begin{figure*}
	\centering
	\includegraphics[width=0.9\linewidth]{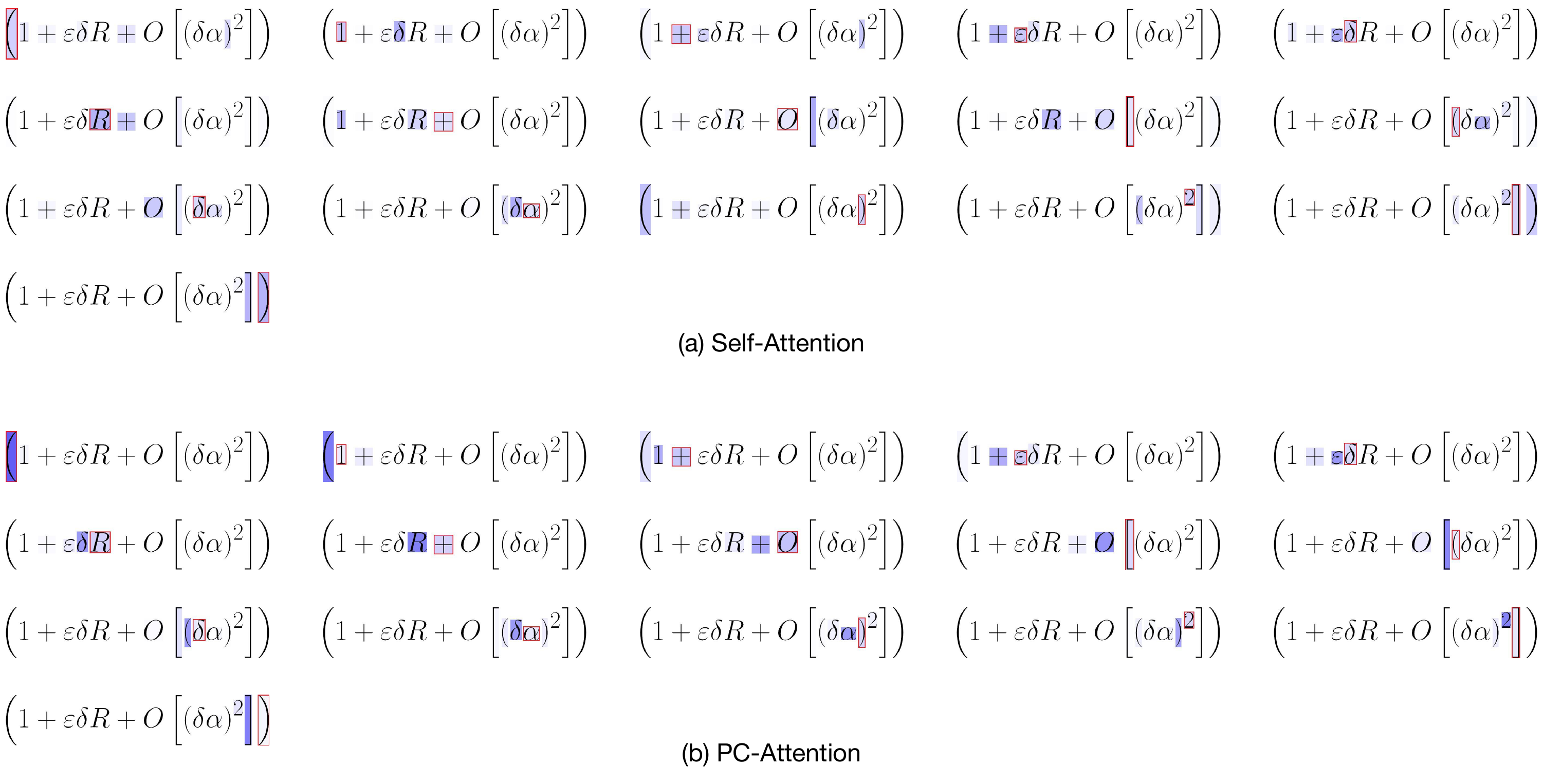}
	\caption{Visualizing different attention mechanisms in reconstruction module.}
	\label{fig:pcEncoder}
\end{figure*}

As demonstrated in Figure~\ref{fig:pcEncoder}, we further visualized the differences between attention mechanisms used in the reconstruction module of encoder, where the target symbol is in a red box. 
For each target symbol, we compare two attention mechanisms to address how they capture the spatial relationships between symbols in the symbol-level image encoder.
From the visualization, we observe that PC-attention focuses on the nearest neighbors of the target symbol.
For every target symbol, the found dependent symbols are reasonable in Figure~\ref{fig:pcEncoder}(b).
However, it is hard to explain the self-attention mechanism, e.g., Columns 2-3 at Line 1, Columns 2, 4 at Line 2, and Columns 1, 3 at Line 3 in Figure~\ref{fig:pcEncoder}(a).
Thus, we can conclude that PC-attention is more reasonable to recover the spatial relationships between symbols in the encoder.

\section{Conclusion}
In this paper, we propose an encoder-decoder framework with symbol-level features to address the PMER problem.
Compared with existing PMER method, the designed symbol-level image encoder aims to preserve the fine-grained symbol features and their spatial information.
For recovering the spatial relationships between symbols, we propose the PC-attention mechanism to restore them in the reconstruction module of encoder.
We have conducted extensive experiments on two real datasets to illustrate the effectiveness and rationality of our proposed EDSL method.

In this work, we have only addressed the PMER problem. 
Thus, it may fail to recognize the handwritten math expressions since they are non-standard compared to printed ones. 
To address this issue, we plan to extend our proposed EDSL method to address the handwritten MER problem. 
In addition, the math expressions are rich in the structural information. 
In the real-world, there are many other images, which contain the structural information, such as music, Chemical equations, Chemical molecular formula, and so on.
Thus, we plan to investigate how to effectively recognize such structural information from images.

\bibliographystyle{unsrt}  
\bibliography{sample-base}

\end{document}